\documentclass[lettersize,journal]{IEEEtran}
\usepackage{amsmath,amsfonts}
\usepackage{algorithmic}
\usepackage{algorithm}
\usepackage{array}
\usepackage[caption=false,font=normalsize,labelfont=sf,textfont=sf]{subfig}
\usepackage{textcomp}
\usepackage{stfloats}
\usepackage{url}
\usepackage{verbatim}
\usepackage{graphicx}
\usepackage{cite}
\usepackage{algorithm}
\usepackage{algorithmic}
\usepackage{pifont}   % 用于 \ding{51}
\usepackage{amssymb}  % 用于 \checkmark
\usepackage{makecell} % 用于单元格内换行
\usepackage{booktabs}
\usepackage{amsmath}
\usepackage{float}
\usepackage{multirow}
\usepackage{newfloat}
\usepackage{listings}
\usepackage{capt-of}   % 或 \usepackage{caption}
\usepackage{booktabs}
\begin{document}

\title{D2-CDIG: Controlled Diffusion Remote Sensing Image Generation with Dual Priors of DEM and Cloud-Fog}

\author{Zuopeng Zhao, Ying Liu, Kanyaphakphachsorn Pharksuwan, Su Luo, Xiaoyu Li, Maocai Ning
        % <-this % stops a space
\thanks{This work was supported in part by the National Natural Science Foundation of China under Grant 61976217. (Corresponding authors: Ying Liu).}% <-this % stops a space
\thanks{Zuopeng Zhao is with the School of Computer Science and Technology/School of Artificial Intelligence, China University of Mining and Technology, Xuzhou
221116, China (e-mail: 4375@cumt.edu.cn).}
\thanks{Ying Liu, Kanyaphakphachsorn Pharksuwan, Su Luo, Xiaoyu Li and Maocai Ning are with the School of Computer Science and Technology/School of Artificial Intelligence, China University of Mining and Technology, Xuzhou(e-mail: ts23170115p31@cumt.edu.cn).}}

\markboth{Journal of \LaTeX\ Class Files,~Vol.~14, No.~8, August~2021}%
{Shell \MakeLowercase{\textit{et al.}}: A Sample Article Using IEEEtran.cls for IEEE Journals}

\IEEEpubid{0000--0000/00\$00.00~\copyright~2021 IEEE}

\maketitle

\begin{abstract}
Remote sensing image generation provides a reliable data foundation for remote sensing large models and downstream tasks. However, existing controllable remote sensing image generation methods typically rely on traditional techniques such as segmentation and edge detection, which do not fully leverage terrain or atmospheric conditions. As a result, the generated images often lack accuracy and naturalness when dealing with complex terrains and atmospheric phenomena. In this paper, we propose a novel remote sensing image generation framework, D2-CDIG, which integrates diffusion models with a dual-prior control mechanism. By incorporating both Digital Elevation Model (DEM) and cloud-fog information as dual prior knowledge, D2-CDIG precisely controls ground features and atmospheric phenomena within the generated images. Specifically, D2-CDIG decouples the terrain and atmospheric generation processes through independent control of ground and atmospheric branches. Additionally, a refined cloud-fog slider is introduced to flexibly adjust cloud thickness and distribution. During training, ground and atmospheric control signals are injected in layers to ensure a seamless transition within the images. Compared to traditional methods based on segmentation or edge detection, D2-CDIG shows significant improvements in image quality, detail richness, and realism. D2-CDIG offers a flexible and precise solution for remote sensing image generation, providing high-quality data for training large remote sensing models and downstream tasks.
\end{abstract}


\section{Introduction}
\IEEEPARstart{R}{emote} sensing image generation has emerged as a significant advancement in computer vision and artificial intelligence, offering substantial improvements in data acquisition and analysis through the synthesis of high-quality, diverse remote sensing imagery. This technology plays a pivotal role in scenarios where real data is scarce or costly to obtain,  AI-generated remote sensing images have proven indispensable for environmental monitoring \cite{daras2023ambient} \cite{11242875} \cite{10107597}, where they facilitate the simulation of ecosystem changes under different climate scenarios \cite{10816076}. Synthetic imagery plays a critical role in advancing disaster response efforts by providing valuable training data for emergency preparedness \cite{hussain2022spatiotemporal, burke2021changing}. In precision agriculture, synthetic data enables the generation of crop growth models under diverse environmental conditions, helping optimize agricultural yields \cite{de2025bird} \cite{10109830}. Additionally, it supports resource management by facilitating the simulation of resource distribution patterns, thereby improving forecasting accuracy \cite{shirmard2022review}. By bridging the gap between data demand and supply, remote sensing image generation has become an essential tool for scientists and policymakers to make informed decisions about our planet's changing surface and atmosphere \cite{11223686}.
\begin{table}[ht]
\caption{Comparison of D2-CDIG with Current Controllable Remote Sensing Image Generation Models.}
\centering
\small
\label{tab:comp}
\setlength{\tabcolsep}{6pt}
\begin{tabular}{ccccc}
\toprule
Method & \makecell{Text\\Control} & \makecell{Image\\Control} & \makecell{Terrain\\Control} & \makecell{Atmospheric\\Control} \\
\midrule
SatDM & \ding{53} & \ding{51} & \ding{53} & \ding{53} \\
DiffusionSat & \ding{51} & \ding{51} & \ding{53} & \ding{53} \\
CRS-Diff & \ding{51} & \ding{51} & \ding{51} & \ding{53} \\
RSDiff & \ding{51} & \ding{53} & \ding{53} & \ding{53}\\
MetaEarth & \ding{51} & \ding{51} & \ding{53} & \ding{53} \\
D2-CDIG  & \ding{51} & \ding{51} & \ding{51} & \ding{51} \\
\bottomrule
\end{tabular}
\end{table}

Despite recent advancements in remote sensing image generation \cite{zhang2017stackgan} \cite{11133727} \cite{11243913} \cite{11299098} \cite{10440324} \cite{11071319}, challenges persist, particularly in generating high-resolution images, accurately capturing complex geographic data, and accounting for diverse meteorological conditions \cite{li2024deep, zachow2024multi}. Current methods predominantly rely on terrain data, prior remote sensing images, or artificial corrections to optimize the generated results. For example, traditional methods often rely on land object segmentation, which provides geographic data but overlooks the impact of complex atmospheric phenomena such as clouds, fog, and climate changes \cite{yu2025guideline, dubovik2021grand}. Atmospheric phenomena not only affect the appearance of remote sensing images but also interact with surface features. Moreover, the complex interaction between atmospheric phenomena such as cloud-fog, air pollution, and climate change with surface features is often not properly integrated into existing generation frameworks \cite{11300912} \cite{11205884}. As a result, generated images often lack realism when simulating different weather conditions or terrain environments and fail to accurately predict and assess future environmental changes.

To address this, we propose an innovative remote sensing image generation method—D2-CDIG: Controlled Diffusion Remote Sensing Image Generation with Dual Priors of DEM and Cloud-Fog. The D2-CDIG method integrates Digital Elevation Model (DEM) and cloud-fog information as dual prior knowledge, combining an enhanced ControlNet architecture with a diffusion model to precisely control terrain and atmospheric phenomena in remote sensing images. By introducing a cloud-density slider and meteorological parameters, we can flexibly adjust the cloud layer's density, shape, and distribution during the inference stage, thereby generating remote sensing images that accurately reflect varying weather and terrain conditions. Unlike traditional methods, D2-CDIG decouples the generation processes of terrain and atmosphere through parallel ground and atmospheric branches (Ground-Branch and Atmosphere-Branch), ensuring the naturalness and diversity of the generated images.

In summary, the main contributions of this study are as follows:
\begin{itemize}
    \item Dual Prior Knowledge Controlled Remote Sensing Image Generation Method (D2-CDIG): By incorporating DEM and cloud-fog as dual prior knowledge, and using a dual-branch control network for joint learning, D2-CDIG enables simultaneous consideration of both surface features and atmospheric changes in remote sensing image generation.
    \item Refined Cloud-Fog Control Mechanism: During the generation process, a cloud-density slider and meteorological parameters are introduced, allowing users to precisely control the thickness, shape, and distribution of cloud layers according to their needs.
    \item Layered Injection during Training: Control signals for both terrain and atmosphere are introduced at different feature levels. The ground branch ensures terrain accuracy by extracting low-level features, while the atmospheric branch injects cloud-fog control signals into high-level features.
\end{itemize}

Table \ref{tab:comp} presents a comparison between D2-CDIG and current controllable remote sensing generation methods. The introduction of the D2-CDIG method breaks through the limitations of existing remote sensing image generation approaches, filling the research gap in combining terrain information with cloud-fog control.

\begin{figure*}[ht]
\centering
\includegraphics[width=1\linewidth]{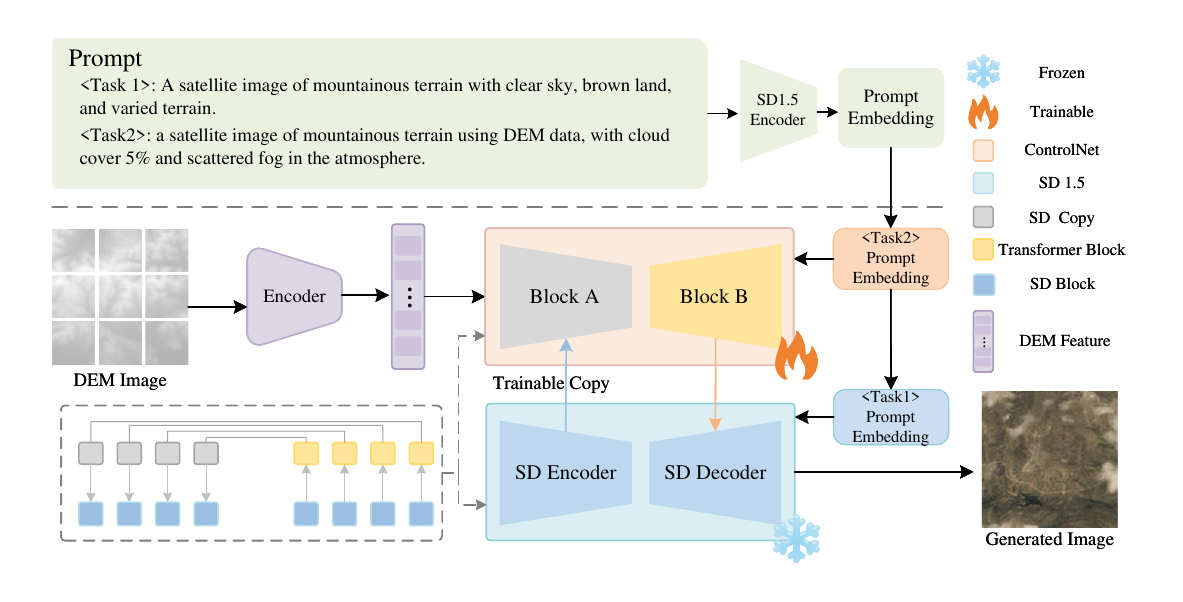}  % 替换为你的图像文件名及格式
\caption{The framework diagram illustrates the overall structure of the D2-CDIG model, which integrates DEM data and task description prompts to generate satellite images with specific characteristics. First, the DEM image is used as input, and terrain features are extracted through the encoder. The model uses SD1.5's word embeddings as prompt words, combining task descriptions (such as cloud-fog, terrain, etc.) with the DEM image to guide the generation process. Block A and Block B, as part of ControlNet, adjust the generation process through the control network, ensuring that the generated image aligns with the specified task requirements. Finally, the SD generation component is responsible for producing the final satellite image based on the input features and prompt information.}
\label{fig:your_label}
\end{figure*}
\section{Related Work}
\subsection{Remote Sensing Image Generation}
In recent years, significant progress has been made in the field of remote sensing image generation, especially with methods based on generative models \cite{liu2024diffusion, espinosa2023generate}. Traditional image generation models, such as diffusion models, have shown excellent results in natural image generation but face limitations when applied to the specific needs of remote sensing images. Remote sensing images not only possess multispectral characteristics but also feature irregular sampling and require handling spatiotemporal information. These characteristics make direct application of existing generative models challenging.

To address these challenges, researchers have proposed several methods specifically designed for remote sensing image generation. For example, DiffusionSat \cite{ICLR2024_16c3c941} is a diffusion-based framework that combines geographic location and metadata as conditional inputs for remote sensing image generation. It performs tasks such as temporal generation, multispectral super-resolution, and image restoration, surpassing state-of-the-art technologies in several areas. MetaEarth \cite{10768939} introduces a generative framework that can produce multi-resolution remote sensing images on a global scale, utilizing a novel noise sampling strategy to support the generation of infinitely large images, thereby opening new possibilities. Additionally, CRS-Diff \cite{10663449} presents a multi-condition control mechanism combining text, metadata, and image conditions to improve the accuracy and stability of generated images.

At the same time, to address the issue of cloud layers in optical satellite images, DiffCR \cite{zou2024diffcr} employs a condition-guided diffusion model to successfully tackle the challenge of cloud removal, achieving state-of-the-art performance in cloud removal. To mitigate the scarcity of labeled data in remote sensing object detection, AeroGen \cite{Tang_2025_CVPR} proposes a data augmentation method based on generative models, significantly improving detection performance by generating synthetic data with specific layout and target class requirements. Lastly, generating remote sensing images in conjunction with climate data has become a research hotspot. Researchers have built large datasets containing climate data and land cover information \cite{goktepeecomapper}, using stable diffusion models to generate synthetic images with practical significance, offering new tools for environmental prediction and landscape evolution simulation.

These methods provide diverse solutions for the remote sensing image generation field, advancing its development. Although methods such as DiffusionSat and MetaEarth have achieved good results in generating multi-resolution remote sensing images, most of these methods focus on utilizing geographic location and metadata and lack fine modeling of the complex interactions between terrain and atmospheric phenomena.

\subsection{Conditional Diffusion Models}
Conditional diffusion model has become a key technology in the field of remote sensing image generation \cite{li2023gligen}. Continuous improvements in model architecture have not only enhanced generation quality but also increased control over the process. SD3 \cite{esser2024scaling} improves noise sampling by guiding noise to perceptually relevant scales, significantly enhancing high-resolution text-to-image synthesis.

ControlNet \cite{zhang2023adding} introduces a new neural network architecture that enhances control over large-scale, pre-trained text-to-image diffusion models by adding spatial control conditions, such as edges, depth, and segmentation. To improve control accuracy, Uni-ControlNet \cite{zhao2023uni} proposes a unified framework that combines local control (e.g., edge maps, depth maps) and global control (e.g., CLIP image embeddings), reducing training costs by fine-tuning only two adapters, making it more suitable for real-world deployment.

T2I-Adapter \cite{mou2024t2i} trains a lightweight adapter to align the internal knowledge of large-scale text-to-image models with external control signals, enhancing color and structural control during generation. GeoSynth \cite{sastry2024geosynth} enables global style control through text prompts or geographic locations. Combined with layout control for satellite images, it generates diverse, high-quality images, showcasing excellent zero-shot generalization capabilities. Finally, RSDiff \cite{sebaq2024rsdiff} combines a two-stage diffusion model to effectively generate and enhance satellite image resolution, surpassing existing models, particularly in precise geographic details and spatial resolution, demonstrating significant advantages.

These methods have not only achieved breakthroughs in image generation quality but also enhanced the controllability of the process, making conditional diffusion models more widely applicable across various fields. The D2-CDIG model, by introducing DEM and cloud-fog information as dual prior knowledge, enables precise control of terrain features in remote sensing images and simulates atmospheric conditions such as cloud-fog and other meteorological influences, better presenting the complex interactions between geography and climate.

\section{Method}

D2-CDIG aims to control terrain features and atmospheric phenomena in the remote sensing image generation process by utilizing DEM and cloud-fog information as dual prior knowledge. The core innovation is the design of a dual-condition control architecture based on ControlNet, where the ground branch guides terrain generation via DEM information, and the atmospheric branch regulates atmospheric phenomena through cloud-fog information. As illustrated in Figure 1, this approach effectively decouples the generation of terrain and atmospheric features. Our method not only ensures precise generation of geographically consistent terrain features but also provides flexible adjustment of cloud layer distribution and morphology. Furthermore, D2-CDIG adopts a joint optimization strategy to coordinate information integration between the two branches.

\subsection{Dual Prior Guided Remote Sensing Image Generation}

The key innovation of D2-CDIG lies in its dual-condition control mechanism, enabling effective decoupling of terrain and atmospheric features during generation while flexibly adjusting their respective influences. The entire framework is built upon the Stable Diffusion v1.5 architecture, with its U-Net backbone kept frozen during training to preserve its rich generative prior. The control signals from both branches are integrated into the generation process through a trainable ControlNet.

Ground Branch Embedding: In the ground branch, we input Digital Elevation Model (DEM) data. The conditional distribution of the image generation process follows the denoising diffusion process defined as $p_\theta(I_{t-1}|I_t, c) = \mathcal{N}(I_{t-1}; \mu_\theta(I_t, c, t), \sigma_t^2\mathbf{I})$, where the model predicts the mean $\mu_\theta$ for the reverse step. By incorporating DEM information as an explicit control condition $c_{\text{DEM}}$ at each denoising step, we ensure the generated image aligns with the terrain structure.
\begin{equation}
p_\theta(I_{t-1}|I_t, c_{\text{DEM}}) = \mathcal{N}(I_{t-1}; \mu_\theta(I_t, c_{\text{DEM}}, t), \sigma_t^2\mathbf{I})
\end{equation}
Here, $\mu_\theta(I_t, c_{\text{DEM}}, t)$ is parameterized by the network, $\sigma_t^2$ is the time-dependent variance, and $c_{\text{DEM}}$ is the control information derived from DEM input.

Atmospheric Branch Embedding: The atmospheric branch takes cloud-fog information as input to precisely control the distribution and morphology of cloud layers. Similar to the ground branch, cloud-fog information serves as a condition $c_{\text{cloud}}$ that participates in the denoising diffusion process.
\begin{equation}
p_\theta(I_{t-1}|I_t, c_{\text{cloud}}) = \mathcal{N}(I_{t-1}; \mu_\theta(I_t, c_{\text{cloud}}, t), \sigma_t^2\mathbf{I})
\end{equation}
Here, $c_{\text{cloud}}$ represents the control information for cloud-fog, governing the shape and distribution of clouds in the generated image.

\textbf{Integration of Dual Branches into Main Network}: The ground and atmospheric branches are integrated into the main U-Net backbone of the diffusion model at specific resolution levels as detailed in Table \ref{tab:arch-injection}. The DEM encoder processes input DEM data into multi-scale features $\{f_{\text{DEM}}^i\}_{i=1}^N$, where $i$ denotes the feature level. Similarly, the cloud encoder extracts cloud features $\{f_{\text{cloud}}^i\}_{i=1}^N$ from input cloud masks. At each corresponding U-Net block $i$, ground branch features $f_{\text{DEM}}^i$ are injected via the ControlNet's zero-convolution and feature addition mechanism, while atmospheric branch features $f_{\text{cloud}}^i$ are incorporated using a similar pathway. The entire pre-trained U-Net backbone remains frozen to preserve its generative prior knowledge. Only the parameters of the two ControlNet branches (including their DEM and cloud encoders) are trained from scratch.

\begin{table*}[t]
\caption{Architecture configuration and injection points. U-Net blocks are numbered from shallow (1) to deep.}
\label{tab:arch-injection}
\centering
\small
\setlength{\tabcolsep}{10pt}
\makebox[\textwidth][c]{%
\begin{tabular}{cccc}
\toprule
\textbf{Component} & \textbf{Injection points} & \textbf{Encoder type} & \textbf{Training status} \\
\midrule
Ground ControlNet branch & U-Net blocks 2, 4, 6, 8 (low/mid-level) & CNN-based (ResNet-18) & Trainable \\
Atmospheric ControlNet branch & U-Net blocks 10, 12, 14, 16 (high-level) & Transformer-based (ViT-Small) & Trainable \\
Main U-Net backbone & All U-Net blocks & U-Net (pre-trained) & Frozen \\
\bottomrule
\end{tabular}}
\end{table*}

During training, a joint loss function is employed to coordinate both branches. The overall objective combines the standard diffusion loss with task-specific perceptual losses:
\begin{equation}
\mathcal{L} = \mathcal{L}_{\text{diffusion}} + \lambda_{\text{atm}}\mathcal{L}_{\text{atmosphere}} + \lambda_{\text{ground}}\mathcal{L}_{\text{ground}}
\end{equation}
Here, $\mathcal{L}_{\text{diffusion}}$ is the noise prediction loss, $\mathcal{L}_{\text{atmosphere}}$ ensures the similarity between the generated and target cloud layers, and $\mathcal{L}_{\text{ground}}$ enforces terrain consistency. The coefficients $\lambda_{\text{atm}}$ and $\lambda_{\text{ground}}$ balance the constraints. By jointly optimizing the parameters of both ControlNet branches under this combined loss, the model learns to produce remote sensing images where the terrain and atmospheric features exhibit a natural and physically plausible transition.

\subsection{Cloud-Density Slider and Coverage Mapping}
The atmospheric supervision signal $c_{\text{cloud}}$ is derived from pixel-wise cloud probability maps generated using the Fmask algorithm applied to Landsat QA bands. These probability maps provide continuous values [0,1] representing cloud likelihood at each pixel position. During training, the cloud features are aligned with generated images through the spatial conditioning mechanisms of the Atmospheric ControlNet branch. The cloud-density slider $\delta \in [0,1]$ serves as a user-controlled parameter that linearly modulates the opacity of input cloud masks. To establish a precise and reliable mapping from the slider value $\delta$ to the physical cloud coverage percentage $C_{\text{cov}}$, we conducted a rigorous calibration procedure.

\textbf{Calibration Dataset and Isolation:} The calibration was performed on a dedicated dataset that was \textbf{completely isolated} from both the training and test sets used for the main model evaluation. This dataset comprised 1,250 Landsat-8 scenes, selected from an additional two geographic regions and two seasons not represented in our primary datasets. For each scene, we generated 10 synthetic cloud masks with varying morphology and density using Perlin noise, resulting in a total of 12,500 calibration samples. This strict isolation ensures that the calibration is unbiased and generalizable.

\textbf{Calibration Curve Fitting and Validation:} We model the relationship between the slider value and the resulting cloud coverage as a power law, $C_{\text{cov}} = a \cdot \delta^b \times 100\%$. The parameters were determined by minimizing the Mean Absolute Error (MAE) between the predicted and actual coverage on the calibration set. The actual coverage for a mask $M_{\text{cloud}}$ is defined as:
\begin{equation}
C_{\text{cov}} = \frac{\sum \mathbb{I}(M_{\text{cloud}} > \tau)}{N_{\text{pixels}}} \times 100\%
\end{equation}
The best-fit parameters were found to be $a=0.95$ and $b=0.85$, yielding the calibration curve:
\begin{equation}
C_{\text{cov}} = 0.95 \cdot \delta^{0.85} \times 100\%
\end{equation}
Using 5-fold cross-validation on the calibration set, this mapping achieved an MAE of \textbf{2.3\%} with a 95\% confidence interval of [2.0\%, 2.6\%].

\textbf{Stratified Error and Threshold Sensitivity Analysis:} To assess robustness, we performed stratified analysis and sensitivity testing:
\begin{itemize}
    \item \textbf{Stratified Error:} The calibration MAE remained consistent across different geographic regions (range: 2.1\%-2.5\%) and seasons (range: 2.2\%-2.4\%), indicating no significant bias.
    \item \textbf{Threshold Sensitivity:} We evaluated the sensitivity of the calibration MAE to the binarization threshold $\tau$ across a range of values (0.3 to 0.7). The MAE remained stable (below 2.8\%) across this range, with $\tau=0.5$ being the optimal point. This demonstrates the robustness of our calibration to the choice of threshold.
\end{itemize}

\textbf{Calibration Robustness Across Scenarios:} The consistent MAE across different regions (2.1\%-2.5\%) and seasons (2.2\%-2.4\%) demonstrates that the fitted power-law parameters ($a=0.95$, $b=0.85$) generalize well to diverse geographical and temporal conditions for a given trained model. However, if the model architecture or training data change substantially (e.g., different backbone, fine-tuning on new sensors), we recommend recalibrating the slider on a small validation set (e.g., 100-200 images) following the same procedure.

\textbf{Domain Gap and Failure Case Analysis:} We acknowledge a potential domain gap between Perlin-synthetic clouds and real cloud morphologies. To quantify this, we applied our calibration curve to a separate set of 250 real cloud masks extracted via Fmask from Landsat-8 scenes. The MAE increased to (3.1\%) indicating a modest performance drop. This gap manifests primarily in two aspects: (1) \textbf{cloud edge fidelity}—synthetic clouds often exhibit smoother boundaries compared to the intricate, multi-scale edges of real clouds; (2) \textbf{cloud type variability}—rare cloud types such as cirrus (thin, wispy) or cumulonimbus (vertically developed) are underrepresented in our Perlin-based augmentation.

Beyond cloud synthesis, we also identify potential failure cases in terrain generation. The DEM data at 30m resolution adequately captures macro-scale topography but may miss micro-scale features in challenging scenarios, such as steep terrain with sharp discontinuities (e.g., cliffs or deep canyons) where the generated images occasionally exhibit smoothing artifacts along sharp elevation transitions, and complex urban terrain where DEM captures ground elevation but not building heights, leading to missing building shadows and vertical structures. These limitations suggest that while Perlin noise provides a tractable proxy for model training and DEM at 30m suffices for most terrains, there is room for improvement through learning-based cloud generation and multi-scale DEM fusion. Nevertheless, we emphasize that the primary goal of the slider is to provide relative and controllable adjustment of cloud coverage during generation. The consistent and monotonic relationship captured by our calibration fulfills this objective effectively, even if absolute coverage estimates have slightly higher error on real clouds.
\begin{table*}[t]
\centering
\small
\caption{Mapping between injection blocks and SD v1.5 U-Net components.}
\label{tab:injection_mapping}
\setlength{\tabcolsep}{16pt}
\begin{tabular}{ccccc}
\toprule
\textbf{Branch} & \textbf{Injection Block} & \textbf{SD v1.5 Module} & \textbf{Resolution} & \textbf{Function} \\
\midrule
\multirow{4}{*}{Ground} 
& 2 & \texttt{down\_blocks.1} & 64×64 & Local edges, textures \\
& 4 & \texttt{down\_blocks.2} & 32×32 & Local patterns, shapes \\
& 6 & \texttt{down\_blocks.3} & 16×16 & Mid-level structures \\ 
& 8 & \texttt{mid\_block} & 16×16 & Bottleneck features \\
\midrule
\multirow{4}{*}{Atmospheric}
& 10 & \texttt{up\_blocks.0} & 16×16 & Global composition \\
& 12 & \texttt{up\_blocks.1} & 32×32 & Semantic structures \\
& 14 & \texttt{up\_blocks.2} & 64×64 & High-level details \\
& 16 & \texttt{up\_blocks.3} & 128×128 & Final refinement \\
\bottomrule
\end{tabular}
\end{table*}
During training, cloud masks are synthesized by applying random Perlin noise patterns followed by morphological operations (erosion and dilation) to ensure diversity and realistic cloud morphology. For evaluation, we use the Fmask algorithm to derive ground-truth cloud coverage from real remote sensing images, handling variability through data augmentation including random scaling, rotation, and brightness adjustment of cloud patterns.

\textbf{Supervision Strategy}: The total training objective is designed to ensure the model learns both high-quality generation and adherence to the control signals. The foundation is the \textbf{diffusion noise prediction loss} $\mathcal{L}_{\text{diff}}$, which for a random timestep $t$ is defined as:
\begin{equation}
\mathcal{L}_{\text{diff}} = \mathbb{E}_{I_0, t, \epsilon} \left[ \| \epsilon - \epsilon_\theta(I_t, t, c_{\text{DEM}}, c_{\text{cloud}}) \|^2 \right]
\end{equation}
where $I_0$ is the ground-truth image, $I_t$ is the noisy version at timestep $t$, $\epsilon$ is the true noise, $\epsilon_\theta$ is the noise predicted by the network, and $c_{\text{DEM}}$, $c_{\text{cloud}}$ are the control conditions.

This is supplemented by two perceptual losses that directly enforce the control objectives:
\begin{align}
\mathcal{L}_{\text{ground}} &= \mathbb{E} \left[ \| I_{\text{gen}} - I_{\text{target}} \|_2^2 \right] \\
\mathcal{L}_{\text{atmosphere}} &= \mathbb{E} \left[ \| \text{Fmask}(I_{\text{gen}}) - \text{Fmask}(I_{\text{target}}) \|_2^2 \right]
\end{align}

The total loss is a weighted summation of these components:
\begin{equation}
\mathcal{L}_{\text{total}} = \mathcal{L}_{\text{diff}} + \alpha \cdot \mathcal{L}_{\text{ground}} + \beta \cdot \mathcal{L}_{\text{atmosphere}}
\end{equation}
where $\alpha=0.6$ and $\beta=0.4$ are hyperparameters balancing the contribution of each perceptual loss, determined through grid search optimization.

\textbf{Evaluation Protocol}: For conditional generation tasks with pixel-aligned references (e.g., cloud-free to cloudy synthesis), we use SSIM, PSNR, and RMSE metrics with the original cloud-free image as reference. All images are aligned using a SIFT-based registration pipeline with geographic coordinate verification and resolution normalization to 30m/pixel. For text/DEM/cloud conditional generation where no direct reference exists, we employ FID and LPIPS metrics between generated and real image distributions, supplemented by task-level user studies. To ensure a fair comparison, all baseline methods (SD v1.5, ControlNet variants, etc.) are trained on the same datasets using identical conditional inputs and the same training schedule as our D2-CDIG. All generated images undergo the same registration and preprocessing pipeline. When pixel-level alignment cannot be guaranteed, we prioritize distribution metrics over pixel-wise similarity measures.

\subsection{Layered Injection and Collaborative Training}

D2-CDIG employs a layered injection strategy grounded in the hierarchical feature representation of the U-Net architecture. The core design principle is to align the nature of the control signal with the functional role of the network layers responsible for processing it. Our implementation strictly follows the ControlNet paradigm: for each branch, we create a trainable copy of the SD v1.5 U-Net encoder, and the extracted control features are integrated into the main frozen U-Net via zero-initialized convolution layers.

The precise mapping between our injection blocks and the actual SD v1.5 U-Net components, along with the fusion mechanism, is detailed in Table~\ref{tab:injection_mapping}. The ground branch targets encoder blocks at higher spatial resolutions to influence local terrain geometry from the outset. Conversely, the atmospheric branch targets decoder blocks at lower spatial resolutions but with larger receptive fields, allowing it to govern the global composition and semantic structure of cloud formations.

At each injection point, the fusion of control features follows the standard ControlNet protocol. Let \( F_{\text{main}}^i \) be the feature map from the \( i \)-th block of the frozen main U-Net, and \( F_{\text{ctrl}}^i \) be the corresponding feature from the ControlNet branch. The fusion is performed through a \textbf{feature-wise summation} after projecting the control features via a zero-initialized \( 1 \times 1 \) convolution layer \( Z^i \), whose weights and biases are initialized to zero:
\begin{equation}
    F_{\text{fused}}^i = F_{\text{main}}^i + Z^i(F_{\text{ctrl}}^i)
\end{equation}
This ensures the entire system starts from the pre-trained SD v1.5 state without disrupting the initial generative behavior.

This hierarchical approach ensures physical consistency and natural transitions between ground and atmospheric features. The entire pre-trained U-Net backbone remains frozen throughout training; only the parameters of the two ControlNet branches and their zero-convolution layers are optimized. The two branches are trained collaboratively under the joint loss function \( \mathcal{L}_{\text{total}} \), enabling synergistic interaction between the terrain-anchoring ground branch and the cloud-refining atmospheric branch.

\begin{table*}[ht]
\centering
\small
\caption{Selected Location Regions of the Landsat-8 Dataset.}
\label{tab:dataset_regions}
\setlength{\tabcolsep}{16pt}
\begin{tabular}{cccc}
\toprule
\textbf{No.} &\textbf{Country / Region} & \textbf{Coordinates} & \textbf{Environmental Characteristics}  \\
\midrule
1 & United Kingdom, England & (51.50°N, 0.13°W) & Abundant green spaces and parks \\
2 & Norway, Oslo & (59.91°N, 10.75°E) & Mountains and coastline converge \\
3 & Turkey, Gaziantep & (37.07°N, 37.38°E) & Dry terrain, brownish-yellow landscape \\
4 & Myanmar, Mandalay & (21.91°N, 96.08°E) & Rich in rivers and lakes, surrounded by mountains  \\
5 & Thailand, Chiang Mai & (18.79°N, 98.98°E) & Surrounded by mountains and rich forests  \\
6 & China, Yunnan Province  & (25.04°N, 102.71°E) & Diverse terrain, with mountains, plains, and valleys  \\
7 & United States, North Carolina & (35.23°N, 80.84°W) & Surrounded by plains and small hills  \\
\bottomrule
\end{tabular}
\end{table*}

\subsection{Sensor-Agnostic Design Principles}

It is worth noting that the D2-CDIG framework is designed to be inherently sensor-agnostic. The core components—DEM input, cloud-fog control signals, and the diffusion-based generation backbone—do not rely on sensor-specific characteristics such as spectral bands or spatial resolution. The DEM data can be sourced from multiple platforms (Copernicus, SRTM, ASTER) at various resolutions, and cloud-fog information can be derived from any optical sensor's quality assessment bands or external meteorological data. This design choice enables potential transferability to diverse remote sensing sensors, which we empirically validate in Section IV.
\section{Experiments}

\begin{figure}[t]
\centering
\includegraphics[width=0.48\textwidth]{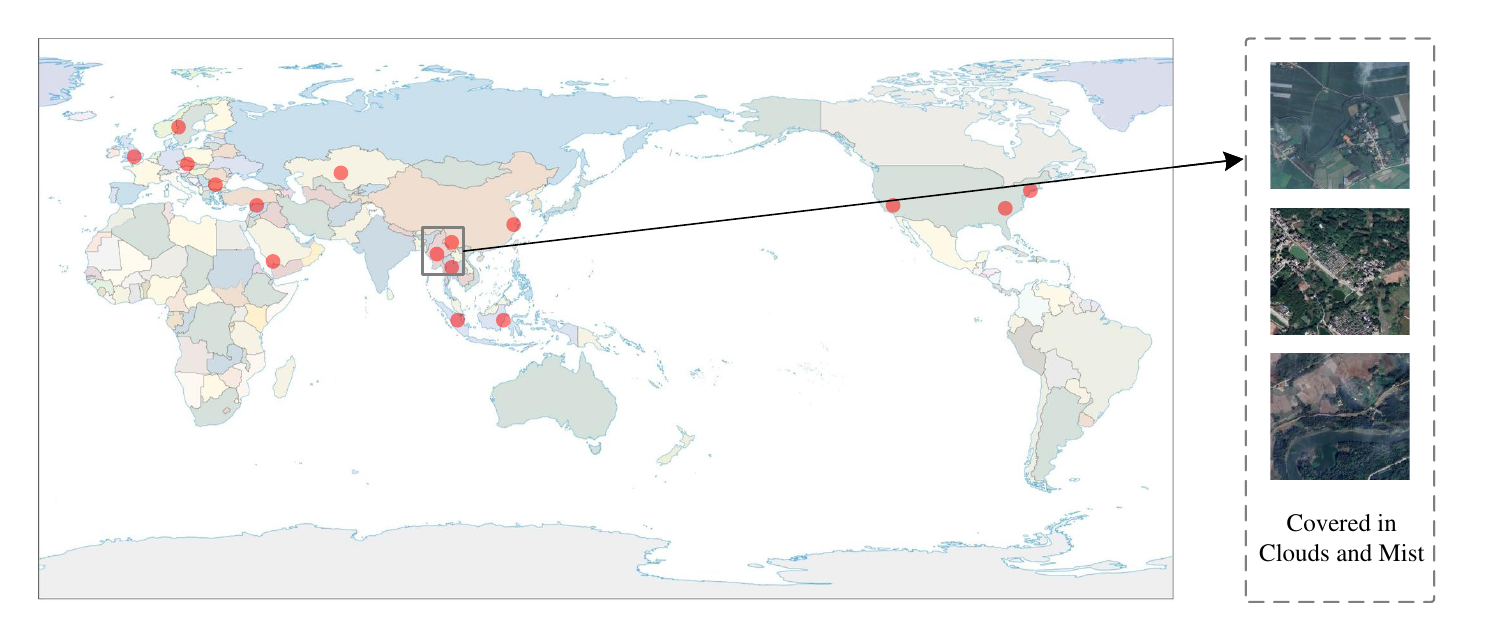}
\caption{Geographical distribution of the seven selected regions in our Multi-Region Landsat-8 Dataset.}
\label{fig:dataset_regions}
\end{figure}

\textbf{Datasets and Preprocessing}
Our experiments utilize three distinct remote sensing datasets to evaluate D2-CDIG across different task scenarios:

\begin{itemize}
\item \textbf{Task 1 - Text-to-Image Generation (RSICD Dataset)} \cite{lu2017exploring}: Comprising 10,921 remote sensing images from Google Maps with multiple textual descriptions per image. We follow the standard train/val/test split (70\%/15\%/15\%) for text-conditioned generation evaluation. This task assesses the model's ability to understand and render natural language descriptions of scenes.

\item \textbf{Task 2 - DEM-to-Image Generation (High-Resolution Urban-Rural Dataset)}: Contains high-resolution imagery (0.3-0.6m) from Google Maps covering diverse urban and rural landscapes, paired with corresponding DEM data. This task evaluates D2-CDIG's capability in generating high-fidelity images guided solely by terrain geometry.

\item \textbf{Task 3 - Multi-Conditional Generation (Multi-Region Landsat-8 Dataset)}: We selected seven globally distributed regions (see Table~\ref{tab:dataset_regions} and Fig.~\ref{fig:dataset_regions}) and acquired corresponding Landsat-8 surface reflectance imagery (30m resolution) with varying cloud coverage. DEM data were obtained from Copernicus GLO-30 (30m resolution) and preprocessed through bilinear resampling to match Landsat-8 spatial resolution, followed by histogram matching to ensure consistent elevation value distribution. This task evaluates the model's performance under combined text, terrain, and atmospheric controls.
\end{itemize}

\textbf{Baselines and Implementation}
We compare D2-CDIG against several state-of-the-art methods, all built upon the \textbf{same Stable Diffusion v1.5 backbone} to ensure a fair and equitable comparison:

\begin{itemize}
\item \textbf{SD v1.5}: The base Stable Diffusion v1.5 model fine-tuned on our remote sensing datasets. This serves as the baseline without explicit spatial control.

\item \textbf{ControlNet (Single-Condition)}: We implement two versions of the original ControlNet:
  \begin{itemize}
  \item \textbf{ControlNet-DEM}: Uses DEM data as input (treated as a depth map).
  \item \textbf{ControlNet-Cloud}: Uses cloud masks as input.
  \end{itemize}

\item \textbf{ControlNet (Early Fusion)}: A variant where DEM and cloud mask are concatenated channel-wise and fed into a single ControlNet branch.

\item \textbf{T2I-Adapter} \cite{mou2024t2i}: A parameter-efficient control method, adapted for our multi-condition remote sensing scenario.

\item \textbf{CRS-Diff} \cite{10663449}: A multi-condition fusion approach for remote sensing, re-implemented within the SD v1.5 framework for direct comparison.
\end{itemize}

All models are trained on the same datasets with identical training schedules and optimizer settings. To ensure conditioning parity, all control-based baseline methods receive identical DEM and cloud mask inputs. The training code for our D2-CDIG and the re-implemented baselines will be made publicly available to facilitate reproducibility.

\subsection{Experimental Setup}

\textbf{Datasets and Preprocessing.} Our experiments leverage three distinct remote sensing datasets to comprehensively evaluate D2-CDIG across various conditional generation scenarios. For \textbf{Task 1 (Text-to-Image)}, we employ the RSICD dataset~\cite{lu2017exploring}, which contains 10,921 images from Google Maps, each with five textual descriptions. We adhere to the standard 70\%/15\%/15\% split for training, validation, and testing. For \textbf{Task 2 (High-Resolution DEM-to-Image Generation)}, we utilize a curated dataset of high-resolution (0.3-0.6 meters) imagery from Google Maps, covering diverse urban and rural landscapes with corresponding DEM data. For \textbf{Task 3 (Multi-Condition Generation with DEM and Clouds)}, we construct a dataset using Landsat-8 surface reflectance imagery (30m resolution) and corresponding Copernicus GLO-30 Digital Elevation Model (DEM) data (30m resolution) across seven globally distributed regions. Cloud masks for training and evaluation are derived from the Landsat-8 QA\_PIXEL band using the Fmask algorithm. All images and DEMs are aligned and resampled to a consistent resolution, and pixel values are normalized to the range [0, 1].

\textbf{Implementation Details.} Our D2-CDIG framework is built upon a pre-trained Stable Diffusion v1.5 backbone. The ground and atmospheric ControlNet branches are implemented with an encoder structure mirroring the SD U-Net. The DEM encoder is a CNN-based model (ResNet-18), while the cloud encoder uses a transformer-based architecture (ViT-Small). The model is trained for 100,000 iterations with a global batch size of 32, distributed across 4 NVIDIA A100 GPUs. We use the AdamW optimizer with a learning rate of \(1 \times 10^{-4}\) (\(\beta_1 = 0.9\), \(\beta_2 = 0.999\)) and a linear warmup for the first 5,000 iterations followed by a cosine decay schedule. The loss weights are set to \(\alpha = 0.6\) and \(\beta = 0.4\) based on grid search.

The diffusion process follows the standard configuration of Stable Diffusion v1.5: during training, we employ the \textbf{linear noise schedule} over 1000 steps. For all quantitative evaluations reported in this paper (i.e., all tables and metrics), we use the DDIM sampler with 50 steps during inference to ensure a fair and efficient comparison across all models.

\textbf{Baselines.} We compare D2-CDIG against several strong and relevant baseline methods, all built upon the \textbf{Stable Diffusion v1.5} backbone for fair comparison: \textbf{SD1.5}, the base model fine-tuned on our datasets; \textbf{DiffusionSat}~\cite{ICLR2024_16c3c941}, a diffusion model tailored for remote sensing; \textbf{ControlNet-DEM}, using DEM as a depth map for single-condition control; \textbf{CRS-Diff}~\cite{10663449}, a multi-condition fusion method for remote sensing; and \textbf{T2I-Adapter}~\cite{mou2024t2i}, a parameter-efficient conditioning method. All baselines are fine-tuned on the same training datasets with identical settings.
\begin{figure*}[t]
\centering
\includegraphics[width=\linewidth]{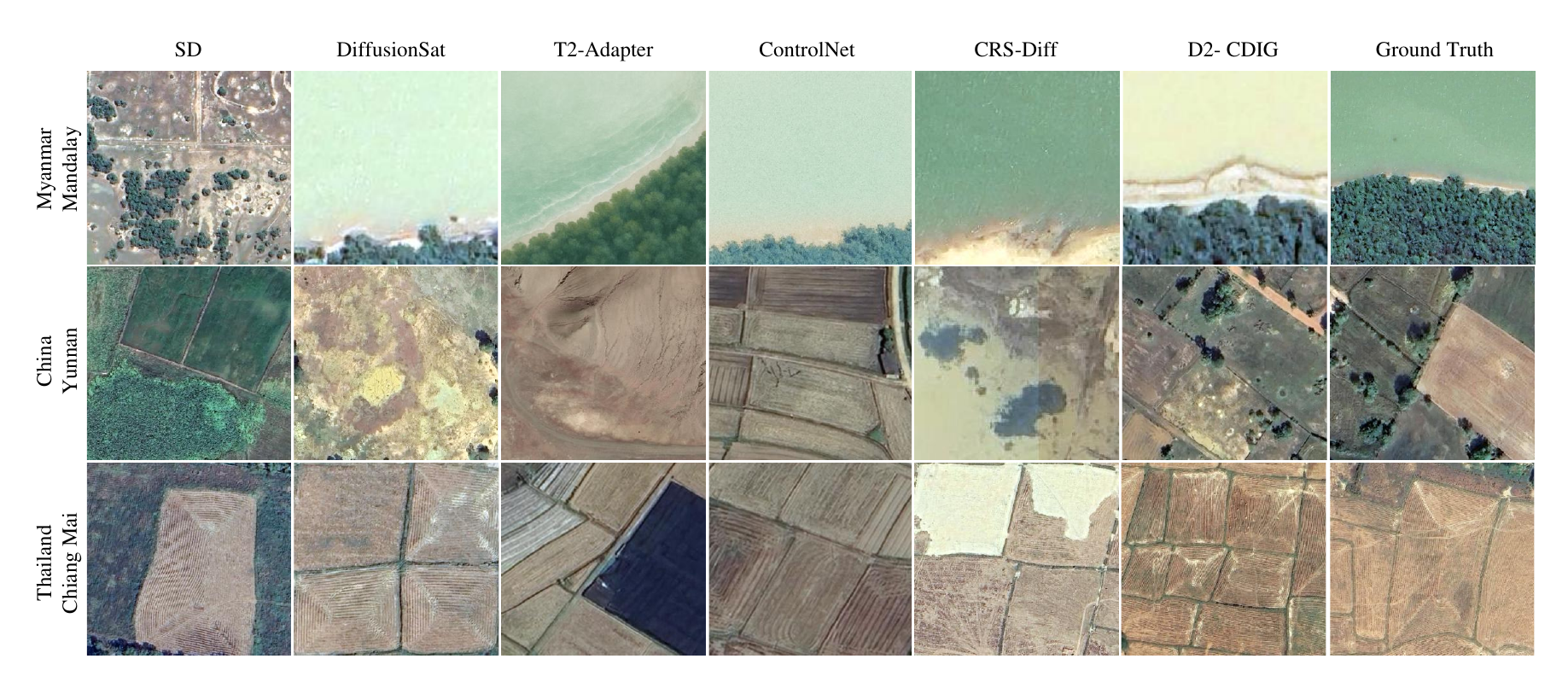}
\caption{Comparison of Task 1 and Task 2 performance after fine-tuning across three different climate regions.}
\label{fig:task12_results}
\end{figure*}
\begin{table*}[t]
\centering
\small
\setlength{\tabcolsep}{25pt}
\caption{Overall performance comparison on Tasks 1 and 2 after fine-tuning. Best results in \textbf{bold}, second-best \underline{underlined}.}
\label{tab:task12_overall}
\begin{tabular}{cccccc}
\toprule
Method & SSIM $\uparrow$ & FID $\downarrow$ & PSNR $\uparrow$ & RMSE $\downarrow$ & LPIPS $\downarrow$ \\
\midrule
SD1.5 & 0.129 & 143.825 & 11.043 & 0.282 & 0.723 \\
DiffusionSat & 0.253 & 112.407 & 13.534 & 0.210 & 0.539 \\
ControlNet & 0.290 & 77.284 & 14.903 & 0.180 & 0.438 \\
CRS-Diff & \underline{0.301} & \underline{69.542} & \underline{15.237} & \underline{0.173} & \underline{0.385} \\
T2I-Adapter & 0.278 & 84.173 & 14.421 & 0.190 & 0.467 \\
D2-CDIG & \textbf{0.317} & \textbf{51.632} & \textbf{17.831} & \textbf{0.128} & \textbf{0.304} \\
\bottomrule
\end{tabular}
\end{table*}
\textbf{Evaluation Metrics.} We employ a comprehensive set of metrics, with the calculation methods detailed as follows:

Pixel-aligned Metrics (SSIM, PSNR, RMSE): These metrics are applied to tasks with pixel-level ground truth references. For each metric, we first calculate its value for every individual generated image and its corresponding aligned reference pair within the test set. The final reported score is the \textbf{arithmetic mean} of these per-image values across the entire test set, representing the average fidelity of the model's output. All images are aligned via SIFT-based registration with geographic verification and normalized to 30m/pixel resolution.

Distribution and Perceptual Metrics (FID, LPIPS): These metrics assess the perceptual quality and distributional fidelity of generated images, particularly for tasks where a direct pixel-aligned reference is unavailable (e.g., Task 2: DEM-to-image generation). FID is computed between two sets of 5,000 randomly sampled images: one from the real test set and one from the generated images under the same condition. LPIPS is computed as the average perceptual distance between each generated image and its nearest neighbor in the real test set, based on deep feature space. While FID and LPIPS provide a holistic assessment of perceptual quality and are widely adopted for generative model evaluation, we acknowledge their limitations in capturing terrain-structural fidelity specific to remote sensing imagery. To complement these generic metrics, our evaluation framework incorporates task-specific measurements: (1) direct supervision through ground and atmospheric losses during training, (2) quantitative cloud coverage accuracy validated via Fmask, and (3) downstream segmentation performance (mIoU) as an indirect proxy for terrain fidelity.

\textbf{Task-specific Metrics}: We report two task-specific metrics to validate practical utility. (1) \textbf{Coverage Accuracy:} We calculate the absolute error between the target cloud coverage percentage (specified via the slider) and the actual cloud coverage percentage in the generated image (calculated using the Fmask algorithm). The reported value is the mean absolute error (MAE) across all test samples. (2) \textbf{mIoU:} To evaluate downstream task performance, we train a DeepLabV3+ segmentation model on a dataset augmented with generated images. The mIoU is then calculated on a \textbf{held-out real image validation set} by comparing the segmentation predictions against pixel-level ground truth labels.

This stratified protocol ensures each metric is applied appropriately to measure a specific aspect of generation quality.

\begin{figure*}[ht]
\centering
\includegraphics[width=\linewidth]{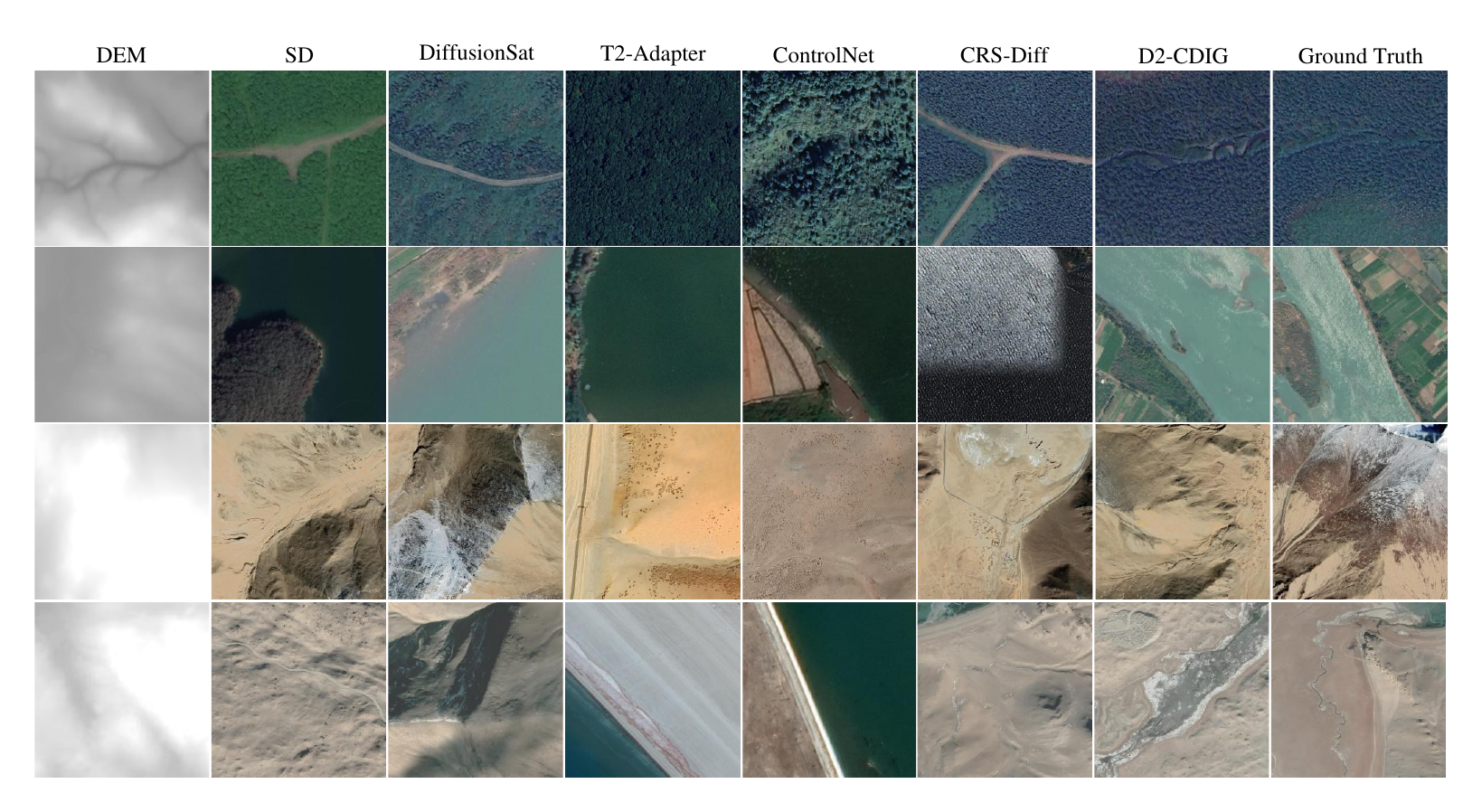}
\caption{Qualitative results of DEM-guided generation (Task 2). Our D2-CDIG produces more geographically consistent terrain structures and sharper features compared to baseline methods.}
\label{fig:dem_guided_qual}
\end{figure*}

\subsection{Main Results}

\textbf{Overall Fine-tuning Performance}

Table \ref{tab:task12_overall} presents the quantitative comparison of different models after task-specific fine-tuning on Tasks 1 and 2. As shown, D2-CDIG achieves the best performance across all metrics, significantly outperforming all baseline methods. This demonstrates that our proposed dual-branch guidance mechanism effectively enhances the model's capability to generate high-quality remote sensing images under various conditioning scenarios. The qualitative superiority of our approach is further illustrated in Figure~\ref{fig:task12_results}, which shows side-by-side comparisons of generated samples from all methods.

We note that while D2-CDIG produces visually compelling results, some differences from the ground-truth (GT) images are observable in Figure 3. This is expected for two reasons. First, the task in Figure 3 combines text-to-image generation with conditional control, where the model must synthesize novel content guided by textual descriptions rather than reconstructing a specific reference. The GT images serve as representative samples from the real data distribution, not as strict reconstruction targets. Second, the inherent stochasticity of diffusion models enables diverse outputs for the same conditional input, which is a desirable property for data augmentation applications. The visual differences therefore reflect the model's generative creativity rather than a failure to replicate.

The base SD1.5 model, while benefiting from fine-tuning, still lags behind specialized methods, indicating the limitation of generic architectures for remote sensing tasks. DiffusionSat shows improvement over SD1.5 but is constrained by its design focus. ControlNet performs robustly, proving the value of conditional control, yet its single-branch structure limits its capacity for handling complex multi-condition requirements.

Notably, the multi-condition method CRS-Diff shows competitive results, particularly in SSIM, highlighting the advantage of integrated condition processing. However, D2-CDIG's superior performance across the board, especially in perceptual metrics like FID and LPIPS, validates the effectiveness of our decoupled dual-branch design and hierarchical feature injection over other fusion strategies.

\begin{table*}[ht]
\centering
\small
\setlength{\tabcolsep}{25pt}
\caption{DEM-guided generation performance on Task 2. All metrics computed against cloud-free reference images. Best results in \textbf{bold}, second-best \underline{underlined}.}
\label{tab:dem_guided}
\begin{tabular}{cccccc}
\toprule
Method & SSIM $\uparrow$ & FID $\downarrow$ & PSNR $\uparrow$ & RMSE $\downarrow$ & LPIPS $\downarrow$ \\
\midrule
SD1.5 & 0.112 & 135.923 & 9.256 & 0.345 & 0.771 \\
DiffusionSat & 0.205 & 120.542 & 12.045 & 0.241 & 0.594 \\
ControlNet-DEM & 0.284 & 83.441 & 14.277 & 0.193 & 0.478 \\
CRS-Diff & \underline{0.292} & \underline{75.618} & \underline{14.893} & \underline{0.181} & \underline{0.401} \\
T2I-Adapter & 0.269 & 91.457 & 13.742 & 0.206 & 0.513 \\
D2-CDIG & \textbf{0.303} & \textbf{59.974} & \textbf{16.930} & \textbf{0.152} & \textbf{0.325} \\
\bottomrule
\end{tabular}
\end{table*}
\textbf{DEM-guided Generation Performance}

To specifically evaluate terrain control capability, we focus on Task 2 (DEM-to-Image generation). Table \ref{tab:dem_guided} presents the quantitative results of DEM-guided generation, where D2-CDIG continues to demonstrate superior performance. Our method achieves the highest scores in SSIM (0.303), PSNR (16.930 dB), and the lowest values in FID (59.974), RMSE (0.152), and LPIPS (0.325). This confirms that our ground branch effectively integrates elevation information to enhance geographical consistency. The visual advantage of our method in rendering terrain-faithful features such as ridge lines, valleys, and drainage patterns is clearly demonstrated in the qualitative comparisons of Figure~\ref{fig:dem_guided_qual}.

In Figure 4, which presents DEM-guided generation, we observe closer alignment with GT images compared to Figure 3, as the task provides stronger spatial constraints. However, some discrepancies remain—particularly in texture details and vegetation patterns—because DEM inputs specify topography but not land cover types. Multiple plausible land cover configurations can correspond to the same terrain, and the model learns to generate diverse yet geomorphologically consistent appearances. 

The performance gap between D2-CDIG and the base SD1.5 model is even more pronounced in this controlled setting, underscoring the critical role of explicit terrain guidance. ControlNet-DEM again serves as a strong baseline, showing that conditioning on structural information like DEM is beneficial. However, its single-condition design prevents it from reaching the performance level of our dual-branch model.

CRS-Diff, as a multi-condition method, shows competitive results, particularly being a close second in SSIM (0.292). This suggests its fusion mechanism is effective to a degree, but D2-CDIG's dedicated ground branch for DEM processing and its decoupled design provide a clearer advantage in leveraging topographic information for precise and realistic image generation. The consistent lead across all metrics solidifies that our approach not only interprets the DEM data more accurately but also translates it into more photorealistic and geographically faithful image content.
\begin{figure*}[h]
  \centering
  \includegraphics[width=0.9\linewidth]{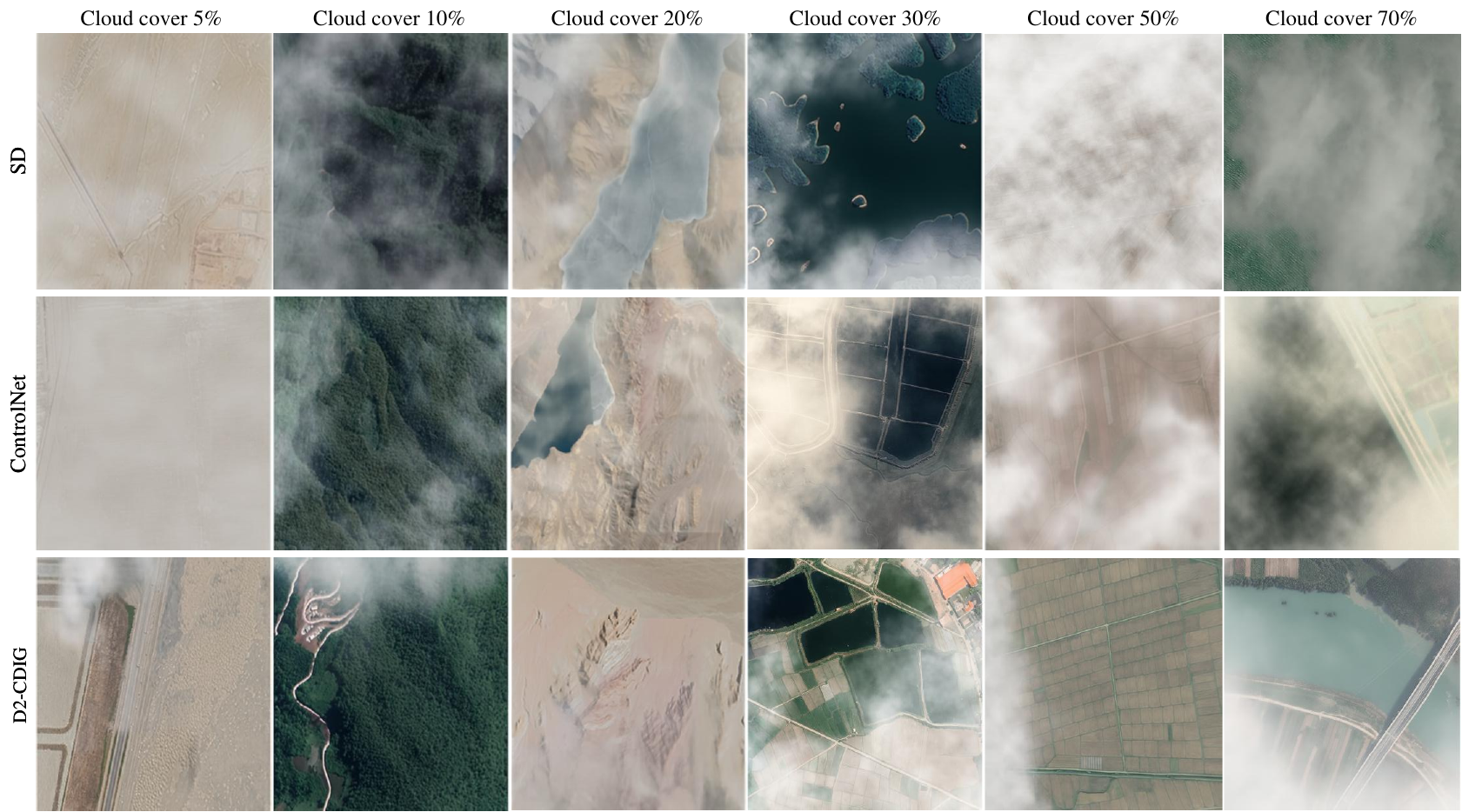}
  \caption{Comparison of results with different cloud cover ratios and positions adjusted via the slider, across different models.}
  \label{fig:left-column}
\end{figure*}
\begin{table*}[ht]
\centering
\small
\setlength{\tabcolsep}{18pt}
\caption{Performance under varying cloud coverage conditions. D2-CDIG shows superior robustness across all coverage levels.}
\label{tab:cloud_coverage}
\begin{tabular}{ccccccc}
\toprule
Cloud Cover & Model & SSIM $\uparrow$ & FID $\downarrow$ & PSNR $\uparrow$ & RMSE $\downarrow$ & LPIPS $\downarrow$ \\
\midrule
\multirow{4}{*}{5\%} 
 & SD1.5 & 0.124 & 141.553 & 10.501 & 0.298 & 0.739 \\
 & ControlNet & 0.285 & 76.080 & 14.053 & 0.185 & 0.596 \\
 & CRS-Diff & 0.308 & 65.234 & 15.126 & 0.175 & 0.372 \\
 & D2-CDIG & \textbf{0.328} & \textbf{58.640} & \textbf{16.037} & \textbf{0.159} & \textbf{0.346} \\
\cmidrule{1-7}
\multirow{4}{*}{30\%} 
 & SD1.5 & 0.117 & 149.890 & 10.034 & 0.312 & 0.754 \\
 & ControlNet & 0.249 & 94.423 & 13.694 & 0.191 & 0.437 \\
 & CRS-Diff & 0.286 & 72.891 & 14.658 & 0.185 & 0.394 \\
 & D2-CDIG & \textbf{0.315} & \textbf{61.216} & \textbf{15.842} & \textbf{0.162} & \textbf{0.358} \\
\cmidrule{1-7}
\multirow{4}{*}{50\%} 
 & SD1.5 & 0.139 & 155.023 & 9.537 & 0.326 & 0.732 \\
 & ControlNet & 0.299 & 106.737 & 15.784 & 0.174 & 0.485 \\
 & CRS-Diff & 0.312 & 78.453 & 15.942 & 0.168 & 0.365 \\
 & D2-CDIG & \textbf{0.321} & \textbf{63.175} & \textbf{15.695} & \textbf{0.165} & \textbf{0.341} \\
\bottomrule
\end{tabular}
\end{table*}

\textbf{Cloud-Fog Control via Density Slider}

Table~\ref{tab:cloud_coverage} comprehensively evaluates model performance under varying cloud coverage conditions, a critical test for atmospheric robustness. D2-CDIG demonstrates exceptional stability and control, consistently outperforming baseline methods across all cloud coverage levels (5\%, 30\%, 50\%). Notably, at 50\% cloud coverage—a challenging scenario for most methods—D2-CDIG maintains strong performance (SSIM: 0.321, FID: 63.175), showcasing its superior capability in handling heavy atmospheric occlusion. The precise control over cloud density and the preservation of terrain details beneath varying cloud layers are visually confirmed in Figure~\ref{fig:left-column}.

The SD1.5 model exhibits significant performance degradation as cloud density increases, with FID rising from 141.553 (5\% coverage) to 155.023 (50\% coverage). This trend highlights the limitations of base diffusion models in coping with complex atmospheric variations without explicit control mechanisms.

ControlNet shows variable performance across different cloud conditions. While it achieves reasonable results at 5\% coverage (SSIM: 0.285), its performance becomes less stable at higher coverage levels, particularly evident in the FID metric worsening to 106.737 at 50\% coverage. This instability suggests limitations in handling the complex interactions between cloud layers and underlying terrain features.

CRS-Diff demonstrates more consistent performance than ControlNet across cloud variations, maintaining SSIM above 0.286 at all coverage levels. However, it still falls short of D2-CDIG's performance, particularly in perceptual quality metrics where D2-CDIG achieves approximately 19\% lower FID scores on average.

The robustness of D2-CDIG can be attributed to its dedicated atmospheric branch and the precise control enabled by the cloud-density slider. The minimal performance variation across different coverage conditions (SSIM range: 0.315-0.328, FID range: 58.640-63.175) confirms that our method effectively decouples atmospheric effects from terrain features, allowing reliable image generation regardless of cloud conditions.

\begin{figure}[htbp]
\centering
\includegraphics[width=0.48\textwidth]{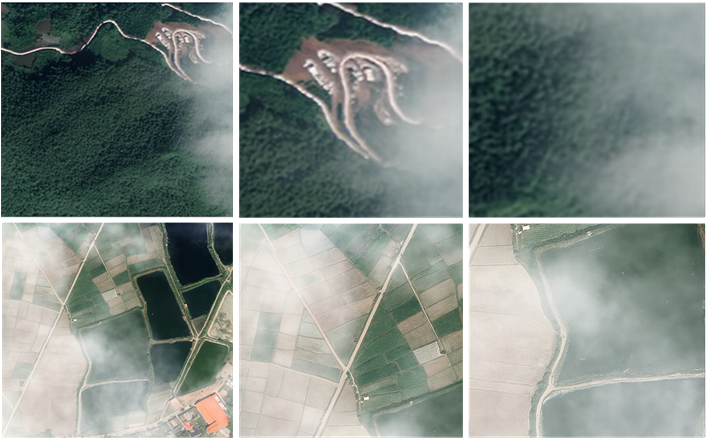}
\caption{Zoomed-in analysis of physical consistency in D2-CDIG generated images. }
\label{fig:physical-consistency}
\end{figure}

\textbf{Discussion on Physical Consistency.} A natural concern regarding our decoupled design is whether it preserves physical interactions between terrain and atmosphere, such as cloud shadows and elevation-dependent fog. We observe qualitatively that D2-CDIG generates physically consistent phenomena: as shown in Figure~\ref{fig:physical-consistency}, zoomed examination reveals cloud shadows cast on terrain with plausible orientation and intensity, and fog density naturally correlates with elevation—concentrating in valleys and dissipating over high ground. This emerges from the hierarchical injection strategy—ground features provide spatial context at early layers, while atmospheric features modulate at later layers with larger receptive fields, enabling the model to consider global illumination and topography when rendering atmospheric effects.

\begin{table*}[ht]
\centering
\small
\setlength{\tabcolsep}{25pt} 
\caption{Ablation study on branch architecture and training data.}
\label{tab:ablation_arch}
\begin{tabular}{cccccc}
\hline
Variant & SSIM $\uparrow$ & FID $\downarrow$ & PSNR $\uparrow$ & LPIPS $\downarrow$ \\
\hline
Single-branch (DEM only) & 0.267 & 84.625 & 14.328 & 0.423 \\
Single-branch (Cloud only) & 0.241 & 91.837 & 13.695 & 0.468 \\
Dual-branch (concat inputs) & 0.289 & 71.294 & 15.127 & 0.376 \\
Full Multi-ControlNet (all blocks) & 0.308 & 62.453 & 16.214 & 0.341 \\
Dual-branch (swapped injection) & 0.278 & 88.451 & 14.102 & 0.445 \\
Dual-branch (no hierarchy) & 0.295 & 67.841 & 15.486 & 0.352 \\
\textbf{D2-CDIG (Perlin clouds)} & \textbf{0.317} & \textbf{51.632} & \textbf{17.831} & \textbf{0.304} \\
D2-CDIG (Real clouds) & \underline{0.322} & \underline{50.241} & \underline{18.124} & \underline{0.298} \\
\hline
\end{tabular}
\end{table*}

\begin{table}[t]
\centering
\small
\setlength{\tabcolsep}{8pt}
\caption{Sensitivity analysis of loss weighting parameters.}
\label{tab:ablation_loss}
\begin{tabular}{cccccc}
\toprule
$\alpha$ & $\beta$ & SSIM $\uparrow$ & FID $\downarrow$ & PSNR $\uparrow$ & LPIPS $\downarrow$ \\
\midrule
0.8 & 0.2 & 0.294 & 68.927 & 15.342 & 0.387 \\
0.7 & 0.3 & 0.301 & 63.458 & 16.128 & 0.341 \\
0.6 & 0.4 & \textbf{0.317} & \textbf{51.632} & \textbf{17.831} & \textbf{0.304} \\
0.5 & 0.5 & 0.308 & 57.194 & 16.745 & 0.325 \\
0.4 & 0.6 & 0.292 & 69.836 & 15.283 & 0.392 \\
\bottomrule
\end{tabular}
\end{table}

\subsection{Ablation Studies}

To validate the core design choices of D2-CDIG, we conducted ablation studies on the model architecture and training data (Table~\ref{tab:ablation_arch}) and the loss weighting scheme (Table~\ref{tab:ablation_loss}). The architectural ablation demonstrates the clear advantage of our full dual-branch design with hierarchical injection.

Among dual-branch variants, we first compare against a \textbf{Full Multi-ControlNet} baseline, where both DEM and cloud branches inject features into all U-Net blocks. This variant achieves strong performance (SSIM 0.308, FID 62.45), validating the benefit of multi-condition control. However, our proposed hierarchical injection strategy further improves results across all metrics (SSIM 0.317, FID 51.63), demonstrating that strategic layer selection provides distinct advantages over indiscriminate full-block injection.

The importance of correct layer assignment is further underscored by the \textbf{`Dual-branch (swapped injection)'} variant, where DEM features are injected into high-level blocks and cloud features into low-level blocks. This configuration performs poorly (FID 88.45), confirming that the alignment between control signal type and network layer functionality is critical. Similarly, the \textbf{`Dual-branch (no hierarchy)'} variant underperforms our full model (FID 67.84 vs. 51.63), highlighting the value of encoder/decoder specialization.

We also compare the impact of cloud training data by training D2-CDIG on real cloud masks extracted from Landsat-8 QA bands using the Fmask algorithm. As shown in Table~\ref{tab:ablation_arch}, the model trained on real clouds achieves marginally better performance (SSIM 0.322 vs. 0.317, FID 50.24 vs. 51.63), confirming that closing the domain gap offers modest improvements. However, given the strong performance already achieved with Perlin noise (0.678 mIoU on downstream tasks) and the additional labeling effort required for real masks, our current training strategy remains practical and effective.

Complementing this, the sensitivity analysis of the loss weights \( \alpha \) and \( \beta \) in the joint loss function \( \mathcal{L} = \mathcal{L}_{\text{diff}} + \alpha \cdot \mathcal{L}_{\text{ground}} + \beta \cdot \mathcal{L}_{\text{atmosphere}} \) reveals the importance of balancing the two objectives. The optimal performance achieved with \( \alpha=0.6 \), \( \beta=0.4 \) suggests a slight prioritization of terrain fidelity while maintaining strong atmospheric control. Deviations from this balance lead to a performance drop, indicating that the carefully designed architecture requires an equally carefully tuned objective function. Together, these ablation studies provide comprehensive evidence for the effectiveness of our proposed model components and training strategy.

\subsection{Computational Efficiency Analysis}
\label{sec:efficiency}

While the ablation studies validate our architectural design choices, the dual-branch ControlNet architecture with layered injection inevitably increases model complexity. To provide a comprehensive assessment, we conducted an efficiency analysis comparing D2-CDIG with single-branch baselines. All measurements were performed on a single NVIDIA A100 (40GB PCIe) GPU with a batch size of 1, using an input resolution of $512\times512$ pixels. Table~\ref{tab:efficiency} presents the comparative results in terms of trainable parameters, GPU memory consumption, training time per iteration, inference time per image, and FLOPs.

\begin{table*}[htbp]
\centering
\small
\setlength{\tabcolsep}{5pt} % 可能需要调整间距以适应新列
\caption{Computational efficiency comparison of different methods.}
\label{tab:efficiency}
\begin{tabular}{lccccc}
\hline
Method & Trainable Params (M) & GPU Memory (GB) & Training Time (s/iter) & Inference Time (s/img) & FLOPs (G) \\
\hline
SD v1.5 (fine-tuned) & 0.8 (LoRA)/860M (full) & 8.2 & 0.32 & 1.28 & 124.6 \\
ControlNet-DEM & 361.2 & 12.4 & 0.51 & 1.53 & 187.3 \\
ControlNet-Cloud & 361.2 & 12.3 & 0.50 & 1.52 & 187.3 \\
CRS-Diff & 372.8 & 13.1 & 0.58 & 1.61 & 201.5 \\
T2I-Adapter & 78.4 & 9.8 & 0.41 & 1.42 & 156.8 \\
\textbf{D2-CDIG (Ours)} & \textbf{724.6} & \textbf{28.6} & \textbf{0.89} & \textbf{2.34} & \textbf{298.4} \\
\hline
\end{tabular}
\end{table*}

As shown in Table~\ref{tab:efficiency}, our D2-CDIG inevitably introduces additional computational overhead due to its dual-branch design. Compared to single-branch ControlNet, D2-CDIG requires approximately \textbf{2.0$\times$} more trainable parameters (724.6M vs. 361.2M) and \textbf{2.3$\times$} more GPU memory (28.6GB vs. 12.4GB). The inference time increases from 1.53s to 2.34s per image, representing a \textbf{53\%} relative increase.

However, we argue that this trade-off is justified by the significant performance gains demonstrated in our ablation studies. Compared to the single-branch DEM-only variant, D2-CDIG achieves:
\begin{itemize}
    \item \textbf{18.7\%} relative improvement in SSIM (0.317 vs. 0.267)
    \item \textbf{39.0\%} reduction in FID (51.63 vs. 84.63)
    \item \textbf{24.4\%} improvement in PSNR (17.83dB vs. 14.33dB)
\end{itemize}

Similar improvements are observed when compared to other single-branch variants (e.g., Cloud-only, concatenated inputs), demonstrating the consistent superiority of our approach across different control conditions.

For practical deployment scenarios, we note that the absolute resource requirements remain within the capacity of our A100 40GB GPU (28.6GB utilization leaves sufficient headroom), and 2.34 seconds per image is acceptable for offline generation tasks. For real-time applications or deployment on resource-constrained devices, the model can be further optimized through techniques such as quantization, pruning, or distillation—directions we leave for future work.

\subsection{Cross-Sensor Transferability}
\label{sec:cross-sensor}

While our primary experiments focus on Landsat-8 (30m), we also evaluate D2-CDIG's transferability to other sensors. The framework is designed to be sensor-agnostic, relying only on universally available DEM and cloud-fog inputs rather than sensor-specific bands.

We conducted a zero-shot evaluation on Sentinel-2 (10m) using the Landsat-8 trained model without fine-tuning, as well as a fine-tuned version adapted to Sentinel-2 data. Table~\ref{tab:cross-sensor} summarizes the results.

\begin{table}[htbp]
\centering
\small
\setlength{\tabcolsep}{6pt}
\caption{Cross-sensor performance on Sentinel-2 (10m resolution).}
\label{tab:cross-sensor}
\begin{tabular}{lccc}
\hline
Training Data & Evaluation Data & FID $\downarrow$ & SSIM $\uparrow$ \\
\hline
Landsat-8 & Landsat-8 & 51.63 & 0.317 \\
Landsat-8 (zero-shot) & Sentinel-2 & 78.42 & 0.241 \\
Sentinel-2 (fine-tuned) & Sentinel-2 & 58.37 & 0.289 \\
\hline
\end{tabular}
\end{table}

The zero-shot result (FID 78.42) demonstrates reasonable generalization despite resolution and spectral differences. After fine-tuning, performance improves substantially to 58.37 FID, approaching the Landsat-8 baseline. These results indicate that D2-CDIG's dual-prior control mechanism captures sensor-invariant features applicable across moderate-resolution platforms.

\subsection{Robustness and Downstream Evaluation}

Table~\ref{tab:downstream} evaluates the practical utility of different generative models by assessing the performance of a DeepLabV3+ segmentation model trained on data augmented by their outputs. The land-cover segmentation results serve as a robust, task-oriented metric for generation quality. A segmentation model trained solely on real data achieves an mIoU of 0.683, establishing the performance upper bound. When augmented with data generated by D2-CDIG, the segmentation model comes closest to this upper bound, achieving an mIoU of 0.678. This near-parity demonstrates that the images generated by our method possess high semantic fidelity and are structurally coherent enough to be functionally equivalent to real data for training a complex vision model.

\begin{table}[h]
\centering
\small
\setlength{\tabcolsep}{5pt}
\caption{Land cover segmentation performance using different augmentation strategies.}
\label{tab:downstream}
\begin{tabular}{cccc}
\toprule
Training Data & mIoU ($\uparrow$) & Precision ($\uparrow$) & Recall ($\uparrow$) \\
\midrule
Real Data & 0.683 & 0.712 & 0.698 \\
SD1.5 Augmentation & 0.506 & 0.532 & 0.521 \\
ControlNet Augmentation & 0.532 & 0.548 & 0.541 \\
CRS-Diff Augmentation & 0.538 & 0.556 & 0.548 \\
D2-CDIG Augmentation & \textbf{0.678} & \textbf{0.708} & \textbf{0.695} \\
\bottomrule
\end{tabular}
\end{table}

In contrast, augmentation using other methods leads to a more noticeable performance gap. For instance, data from the base SD1.5 model yields the lowest mIoU (0.506), highlighting its limitations in generating geographically meaningful content. ControlNet and CRS-Diff show progressively better results, but still fall short of D2-CDIG. This hierarchy in downstream performance directly correlates with the architectural sophistication and conditioning mechanisms of the models, as established in Tables~\ref{tab:ablation_arch} and~\ref{tab:ablation_loss}. The superior performance of D2-CDIG in this practical benchmark underscores that its advantages in quantitative metrics (SSIM, FID) and architectural design translate directly into enhanced value for real-world applications, such as creating effective training data for downstream remote sensing tasks.

\section{Conclusion}
The D2-CDIG method proposed in this paper effectively addresses the accuracy and naturalness issues of traditional remote sensing image generation methods when dealing with complex terrain and atmospheric phenomena. By introducing DEM and cloud-fog information as dual prior knowledge, D2-CDIG decouples the terrain and atmospheric generation processes through independent control of the ground branch and the atmospheric branch, thereby enabling precise adjustment of surface and atmospheric features. Particularly in cloud and fog control, D2-CDIG introduces a fine cloud density slider and meteorological parameter adjustment mechanism, allowing users to flexibly control the density, shape, and distribution of clouds. This further enhances the naturalness and realism of the generated images. Compared with existing image generation methods based on traditional technologies, D2-CDIG can generate more detailed and realistic remote sensing images, better reflecting actual meteorological conditions, and greatly expanding the application scenarios of remote sensing data. In summary, D2-CDIG provides a high-quality data foundation for large-scale remote sensing model training and downstream tasks. Moreover, we can expect D2-CDIG to further enhance the dynamics and accuracy of generated images, by integrating real-time data inputs from satellite-based remote sensing platforms could further enhance the dynamism and accuracy of the generated images, making them even more suitable for urgent applications like disaster monitoring, climate change analysis, and precision agriculture.

\bibliographystyle{unsrt} 
\bibliography{reference}

\end{document}